\documentclass[a4paper,twoside]{article}

\usepackage{graphics} 

\usepackage{mathptmx} 
\usepackage{times} 
\usepackage{mathtools}
\usepackage{multirow}
 
\usepackage[inline]{enumitem}

\makeatletter
\let\NAT@parse\undefined
\makeatother
\usepackage{hyperref}
\hypersetup{
    colorlinks=true,
    linkcolor=blue,
    filecolor=magenta,      
    urlcolor=blue,
    }

\usepackage{todonotes}

\usepackage{epsfig}
\usepackage{epstopdf}
\usepackage{subcaption}
\usepackage{calc}
\usepackage{amssymb}
\usepackage{amstext}
\usepackage{amsmath}
\usepackage{amsthm}
\usepackage{multicol}
\usepackage{pslatex}
\usepackage{apalike}
\usepackage{algorithm2e}
\usepackage[bottom]{footmisc}
\usepackage{SCITEPRESS}     

\begin{document}

\title{NAMOUnc: Navigation Among Movable Obstacles with Decision Making on Uncertainty Interval}

\author{\authorname{Kai ZHANG\sup{1,2,3}\orcidAuthor{0000-0003-1129-9944}, Eric LUCET\sup{1}\orcidAuthor{0000-0002-9702-3473}, Julien ALEXANDRE DIT SANDRETTO\sup{2}\orcidAuthor{0000-0002-6185-2480}, \\ Shoubin CHEN\sup{3,*}\orcidAuthor{0000-0002-9071-0051} and David FILLIAT\sup{2}\orcidAuthor{0000-0002-5739-1618}}
\affiliation{\sup{1}Paris-Saclay University, CEA, List, 91120  Palaiseau, France}
\affiliation{\sup{2}U2IS, ENSTA Paris, Institut Polytechnique de Paris, 91120 Palaiseau, France}
\affiliation{\sup{3}Guangdong Laboratory of Artificial
Intelligence and Digital Economy(SZ), 51800 Shenzhen, China}
\affiliation{\sup{*}Corresponding author}
\email{kaizhangpostbox@gmail.com}
}

\keywords{Navigation among movable obstacles, Planning under uncertainty, Decision making}

\abstract{Navigation among movable obstacles (NAMO) is a critical task in robotics, often challenged by real-world uncertainties such as observation noise, model approximations, action failures, and partial observability. Existing solutions frequently assume ideal conditions, leading to suboptimal or risky decisions. This paper introduces NAMOUnc, a novel framework designed to address these uncertainties by integrating them into the decision-making process. We first estimate them and compare the corresponding time cost intervals for removing and bypassing obstacles, optimizing both the success rate and time efficiency, ensuring safer and more efficient navigation. We validate our method through extensive simulations and real-world experiments, demonstrating significant improvements over existing NAMO frameworks. More details can be found in our website: \url{https://kai-zhang-er.github.io/namo-uncertainty/}}

\onecolumn \maketitle \normalsize \setcounter{footnote}{0} \vfill

\section{\uppercase{Introduction}}
\label{sec:introduction}

In real applications, robots take actions with partial and noisy observation on the environment, and a given action may cause unexpected effect due to uncertainties. If the robot has over confidence on the observation and action, it can lead to some suboptimal decisions, even to dangerous actions resulting in catastrophic effects, like destroying objects in the workspace. It is therefore essential for the robot to recognize the limitation of its observations and actions, and make decisions with awareness of uncertainty and risks.

Recent studies on task and motion planning incorporate uncertainty and update plans based on the observation and action uncertainties. For example~\cite{safronov2020task,pan2022failure} take success rate (SR) into account in a manipulation task: if a grasp action fails, the planner updates its estimation on SR, then replans for an action sequence with a higher predicted SR. Methods such as probabilistic symbolic planning~\cite{silver2021learning} or Bayes optimization~\cite{curtis2024partially} plan with uncertainty but mainly focus on optimizing SR. They often neglect joint optimization with efficiency, which is crucial for navigation tasks.
\begin{figure}[ht]
    \centering
    \includegraphics[width=0.45\textwidth]{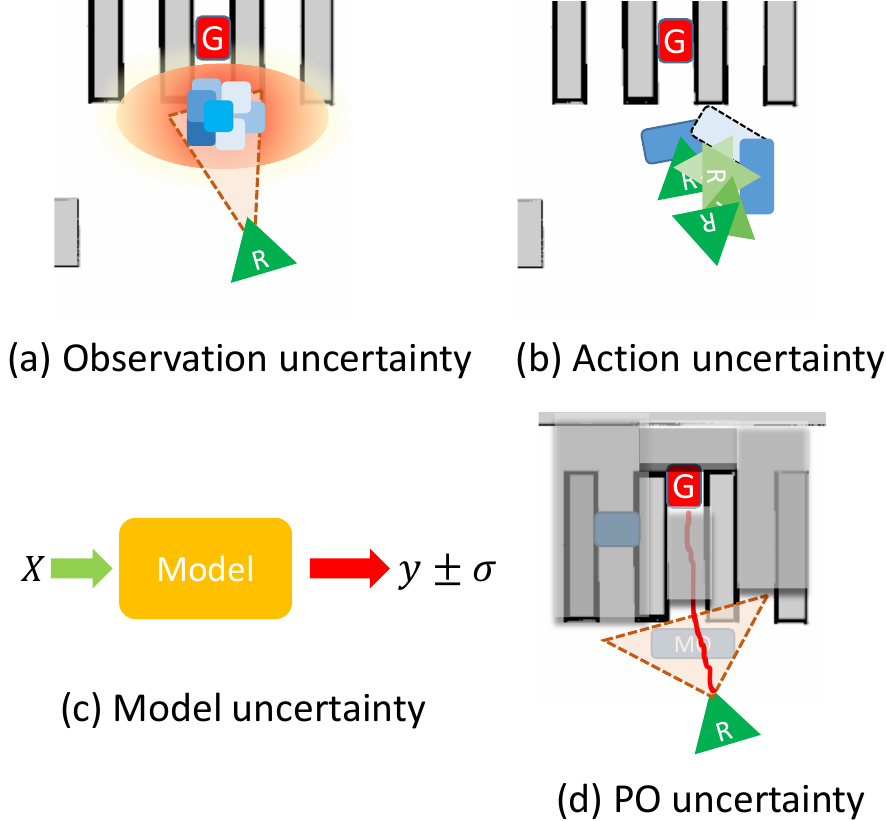}
    \caption{Primary sources of uncertainty in a NAMO task.The illustration highlights four key uncertainties encountered during task execution. The blue rectangle represents the MO while the green triangle with R denotes the robot. The red square labeled G indicates the goal. In subfigure (c), $X$ denotes the observation while $y$ and $\sigma$ represent the prediction result and the corresponding prediction uncertainty, respectively.}
    \label{fig:unc_types}
\end{figure}

Navigation among movable obstacles (NAMO) task is mainly a navigation task but the robot is able to manipulate movable obstacles (MO). Many existing solutions~\cite{muguira2023visibility,ellis2022navigation} consider NAMO as a manipulation task, assuming manipulation is necessary to complete the task. However, the most common case is that the task can be finished without moving the MO, by making a detour. This oversight makes probability-based optimization less useful, as bypassing with a SR close to 1 will be preferred even if it requires significantly more time. Furthermore, in partial observation condition, the invisible region in NAMO tasks is considerably larger than in manipulation tasks. Common strategies for reducing uncertainty in manipulation tasks, such as exploring all occluded regions~\cite{curtis2024partially}, are often inefficient or impractical for NAMO tasks due to the scene complexity and large scale.
%
%

This paper presents NAMOUnc, a method for solving NAMO tasks by optimizing both SR and goal reaching time in real scenarios.
Our main contribution lies in enhancing the robustness of the NAMO method with respect to the following uncertainties (Fig.~\ref{fig:unc_types}): (a) observation uncertainty on MO pose estimation caused by sensor noise, (b) model approximation uncertainty, (c) action uncertainty from imperfect controllers, and (d) blockage uncertainty from partial observation. NAMOUnc estimates these uncertainties as cost intervals and makes decisions based on their utility values, balancing removal and bypass strategies to achieve efficient and successful navigation.


In summary, our contribution includes:
\begin{enumerate}
    \item A novel approach for solving NAMO tasks, optimizing SR and efficiency under condition of partial observability.
    \item Four modules to systematically estimate and quantify the uncertainties described in Fig.~\ref{fig:unc_types}.
    \item A novel method to estimate the uncertainty caused by partial observation in unexplored region, which can effectively reduce the navigation risk and improve the efficiency.
    \item Experiments in simulated and real environments to demonstrate the effectiveness of our method.
\end{enumerate}

\section{\uppercase{Related Work}}
\label{sec:related}
\subsection{Navigation among movable obstacles}

The NAMO task has been extensively studied, with recent advancements focusing on end-to-end learning~\cite{li2020hrl4in} and hybrid approaches~\cite{muguira2023visibility,kim2019learning,xia2021relmogen}. End-to-end learning methods typically employ hierarchical reinforcement learning~\cite{li2020hrl4in}, generating high-level subgoals alongside low-level control parameters. Hybrid methods, on the other hand, leverage machine learning either to produce subgoals~\cite{xia2021relmogen} or to assist in generating action sequences~\cite{muguira2023visibility}. A comprehensive review of these techniques is provided in~\cite{zhang2022TAMP}.

Existing methods primarily focus on task completion without considering the associated costs. A related work~\cite{zhang2023navigation} introduced a strategy selection mechanism based on estimated costs, jointly optimizing SR and efficiency. However, this approach does not incorporate uncertainty, which limits its generalizability and applicability in real-world scenarios. Therefore, we propose NAMOUnc method in this paper as an extension to deal with real-world cases.

\begin{figure*}[ht]
    \centering
    \includegraphics[width=0.95\textwidth]{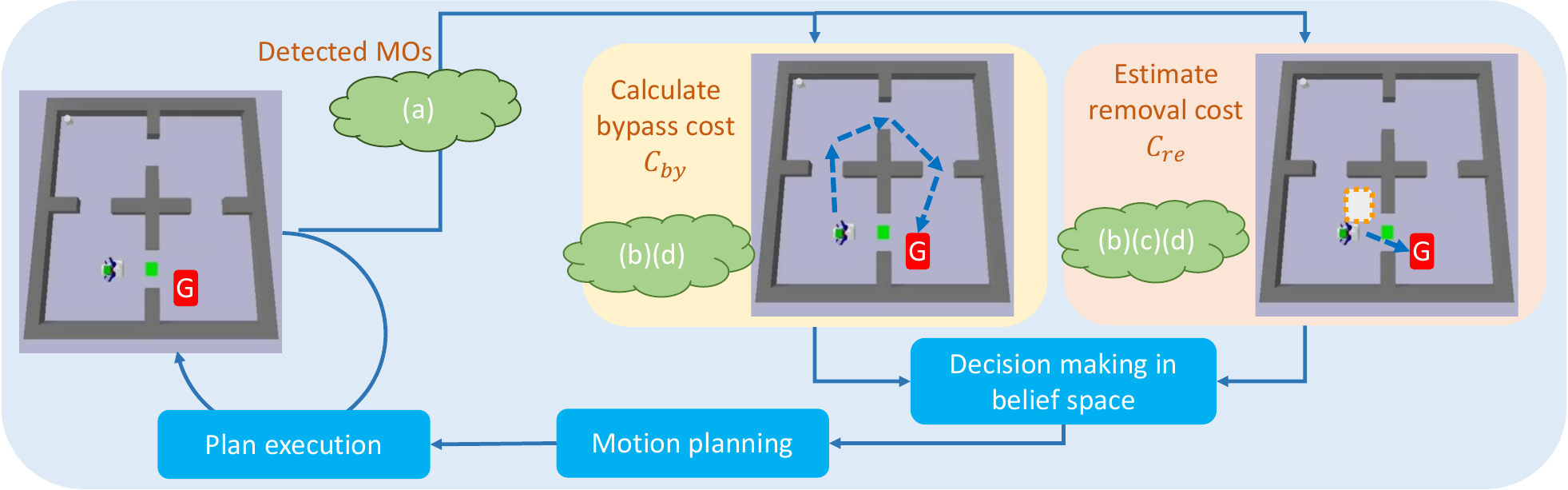}
    \caption{Overview of the NAMO method pipeline. When being blocked by MOs during navigation, the robot estimates the bypass  and removal cost to choose an efficient strategy to continue its task. The green cloud symbols represent the uncertainties associated with each module:~(a) Observation uncertainty;~(b) Model uncertainty;~(c) Action uncertainty;~(d) Blockage uncertainty caused by partial observability.}
    \label{fig:namo_pipeline}
\end{figure*}

\subsection{Planning with uncertainty}

Uncertainty in real-world environments introduces significant challenges, prompting the development of methods to address various types of uncertainties, including observation uncertainty, action uncertainty, and uncertainty arising from partial observability.

\textit{Observation uncertainty} in task planning pertains to the confidence in object classification and pose estimation. Object classification uncertainty has been extensively studied in computer vision~\cite{feng2021review}, with confidence scores commonly used to quantify this uncertainty. Pose uncertainty, often resulting from sensor noise, is often modeled as a Gaussian distribution and mitigated through repeated observations~\cite{kaelbling2013integrated,garrett2020online}.

\textit{Action uncertainty} has garnered increasing attention. In sampling-based methods, action uncertainty is represented as a probabilistic transition matrix, which can be approximated through frequent replanning~\cite{garrett2020online} or learned from demonstrations and datasets~\cite{silver2021learning}. This matrix enables the selection and execution of the most likely successful plan. Rather than treating it as an open-loop process, some approaches~\cite{pan2022failure} iteratively update transition probabilities based on observed action effects, dynamically adjusting plans to achieve task objectives.

Regarding \textit{uncertainty from partial observation}, most methods~\cite{muguira2023visibility,curtis2024partially,wu2010navigation} restrict actions to the sensor's field of view. In manipulation tasks, studies such as~\cite{curtis2024partially,garrett2020online} model the probability of unseen objects behind visible ones, incorporating this into planning to trigger exploration actions when necessary. For NAMO tasks, RNAMO~\cite{wu2010navigation} proposes pushing all obstacles in invisible regions; if obstacles are immovable, the robot updates its internal map and plans a detour. In larger workspaces, LaMB~\cite{muguira2023visibility} employs backward reasoning to eliminate environmental invisibility, though this approach is limited to small environments due to its exhaustive exploration of invisible regions. To address this limitation, we introduce a method to quantify the potential cost of navigating through unexplored areas, optimizing the trade-off between exploration (bypassing into unknown regions) and exploitation (manipulating visible obstacles).

\section{Navigation among movable obstacles with uncertainty}
\label{sec:namo}


In a NAMO task, shown in Fig.~\ref{fig:namo_pipeline}, a robot needs to navigate to a goal while avoiding obstacles. With the environment map, the first step is to plan and follow the shortest path. If meeting a MO blocking this path, the robot can choose to bypass it or clear the path by removing it. As described in our previous work~\cite{zhang2023navigation}, we first estimates the bypass and removal cost before making decision. The bypass cost is calculated based on a detour trajectory while the removal cost is computed in two steps: predicting the stock region for the MO, then estimating the time of moving the MO to this region. The motion planner outputs control parameters for the robot to follow the best alternative.

Uncertainties across different system components can significantly affect task success. In the context of MO detection and localization, observation-related uncertainties, such as recognition errors and pose estimation inaccuracies, can lead to collisions or unsuccessful removal attempts. Similarly, model uncertainties in estimating the costs of bypassing or removing obstacles may yield inaccurate evaluations. If the robot opts to remove a MO, the action may fail due to incomplete knowledge or discrepancies between the predicted and actual outcomes of the action. Furthermore, operating in partially observable environments introduces intrinsic uncertainty, which often results in a divergence between anticipated and actual conditions.

To quantify the impact of these uncertainties on the decision-making process, we use time intervals, denoted as $[\hspace{0.2cm}]$, which reflect the range of potential outcomes generated by uncertainty, before selecting the best option based on their utility values. In the following section, we present the methods of quantifying these uncertainties using intervals.

\section{\uppercase{Uncertainty estimation method}}
\label{sec:method}

We now present four uncertainties considered in NAMO tasks and the methods to estimate them, before detailing the decision process. 

\subsection{Observation uncertainty}
\label{sec:obs_unc}
While navigating the environment, the robot should localize MOs and detect whether the planned path is blocked. Due to the sensors' noise in object detection and localization error of the robot, the estimated MO pose is subject to uncertainty. However, multiple observations can refine the result, and given a specified confidence interval, a belief region representing the MO's position can be calculated for path blockage determination. 

The obstacle pose is computed from the robot pose plus the relative pose of the MO given by the sensor. Assuming the robot pose from the localization algorithm is $X_r=(x_r, y_r, \theta_r)$ with covariance $\Sigma_r$, and the relative distance and angle of the $i$-th MO measured by the depth camera $Y^{i}=(d^i, \phi^i)$ with covariance $\Sigma_Y$, the obstacle pose $X_{MO}^i$ can be obtained by:
$X_{MO}^i=(x_r+d^icos(\theta_r+\phi^i), y_r+d^isin(\theta_r+\phi^i))^T$
and the covariance matrix $\Sigma_{MO}$ by:
\begin{align}
\Sigma_{MO}^i&=J_r \Sigma_r J_r^T+J_y \Sigma_Y J_y^T; \hspace{0,3cm}
J_r=\frac{\partial X_{MO}^i}{\partial X_{r}}; \hspace{0,3cm}
J_y=\frac{\partial X_{MO}^i}{\partial Y^{i}} 
\end{align}

When multiple observations of the same MO are received, a Kalman filter~\cite{kalman1960new} is applied to fuse the repeated observations and obtain the estimated MO pose and its covariance. 

Given a confidence score $T_{conf}$, set to 95\% in our experiments, though other values are possible, the belief region of the MO pose is represented as an ellipse computed from the covariance matrix. We add the size of the MO, expressed as its radius, to obtain the ellipse region where a path would lead to potential collision. In this case, the robot stops and chooses the best strategy as described in Sec.~\ref{Sec:decision}.

\subsection{Bypass cost model uncertainty}
\label{sec:by_predictor}

The bypass cost, defined as the estimated travel time to reach the goal when the robot follows a planned detour, varies depending on the trajectory and the robot's moving speed. A deterministic predictor of navigation time inherently involves uncertainty, since in practice the robot may deviate from the planned trajectory. To capture this prediction uncertainty, we express the estimated time as an interval rather than a single value.

To plan the detour, a MO and its uncertainty region (an ellipse with 95\% confidence as described in Sec.~\ref{sec:obs_unc}) is temporarily added to the obstacle map, then the shortest path planner searches a path to bypass the MO. If no path is found, the bypass time cost is set to  $C^{by}_{nav}=Inf$. 
On the contrary, if a path is found, we need to estimate the navigation time from trajectory features.  While the average speed was simply used in~\cite{zhang2023navigation}, we use a Gaussian linear regressor (GLR)~\cite{williams1995gaussian} for better prediction and uncertainty evaluation (see Sec.~\ref{sec:by_predictor_eval}).

In practice, as rotation takes more time than following straight lines, we need to take the orientation change of the trajectory into account during bypass time estimation. In addition to the trajectory length $F_l$, we therefore calculate the trajectory smoothness $F_{s}$ and variance of direction change $F_v$ as features to estimate the navigation time. Assuming a trajectory consists of $N$ waypoints $pt_i, i=0, 1, .. ,N$, each $pt_i$ characterized by position $p_i$ and orientation $\alpha_i$, smoothness and variance are calculated using: 
\begin{gather} 
\label{deqn_ex1}
 F_s = \frac{\sum_{i=1}^{N} |\alpha_i-\alpha_{i-1}|}{N-1};
F_v = var(|\alpha_i-\alpha_{i-1}|)
\end{gather} 
A trajectory is therefore characterized by $X=\{F_l, F_s, F_v\}$. Then, the bypass time cost and variance are predicted by a GLR:
\begin{equation}
    T_{by},\sigma_{by}=GLR(X)
\end{equation}
The navigation time interval with $T_{conf}=95\%$ confidence is: $[C^{by}_{nav}]=[T_{by} - 2\sigma_{by}, T_{by} + 2\sigma_{by}]$.


To train the regression model, we collect a set of trajectories by controlling the robot to navigate in a warehouse environment. The pose and time stamp are recorded along these trajectories, and we create a large and varied dataset for the model by sampling random start and goal points in these trajectories and computing the corresponding features and duration.

\subsection{Removal action uncertainty}
\label{sec:action_unc}
The action uncertainty relates to the uncertain outcome of loading a MO for displacement, which can be either success or failure.

For a given SR, $p_a$, the expected removal cost $C_{MO}$ is
\begin{multline}
    C_{MO}=T_{MO}\sum^{M}_{i=1}i p_{a}(1-p_{a})^{i-1}+ \\
    (M T_{MO}+C_{by})(1-p_{a})^M
\label{eq:CMO}
\end{multline}
where $M$ defines the maximum attempts when the robot continuously fails to load a MO~(according to the psychological view on learned helpless~\cite{maier1976learned}). $C_{by}$ denotes the obstacle bypass cost, and $T_{MO}$ is the removal cost of a MO that can be estimated by the method proposed in~\cite{zhang2023navigation}, where  a stock region predictor and a regression model are employed to predict the placing pose and the removal time.

In practice, the SR may not be fixed and can evolve during robot operation, we therefore start by an initial estimate and update it after each trial. Similar to~\cite{curtis2024partially}, we model the SR of an action and the uncertainty of its estimation after $t$ trials, $p_a^{t}$, using a Beta distribution. To obtain the initial SR, $p_a^0$, we control the robot to load the MO in several trials~(10 in our experiments) and record the action results. Assuming there are $\alpha$ successful trials and $\beta$ failure cases, the initial knowledge on the trial results can be described as $p_a^0 \sim Beta(\alpha,\beta)$. During operation, when new trials are performed, the updated posterior is $p_a^t \sim Beta(\alpha+s, \beta+f)$ where $s$ and $f$ are the number of successful and failed object manipulation respectively. With a confidence score $T_{conf}$, we can obtain the SR interval: 
\begin{equation}
    p_a^t\in [Beta.ppf(\frac{1-T_{conf}}{2}), Beta.ppf(\frac{1+T_{conf}}{2})]
\end{equation}
 where the $Beta.ppf$ is the point percent function to obtain the confidence interval of the beta distribution given a confidence score. 

We compute the interval of the expected removal cost $[C_{MO}]$ by using Eq.~\ref{eq:CMO} with the minimum and maximum $p_{a}^t$ of the SR interval.

\subsection{Blockage uncertainty}
Blockage uncertainty comes from the partial observation condition and results in a blocking probability of the robot by some unseen  MOs in unexplored region. It is related to the passage width and robot size as it is more risky if the planned navigation path passes a narrow passage rather than an open space. 

\begin{figure}[h]
    \centering
    \includegraphics[width=0.2\textwidth]{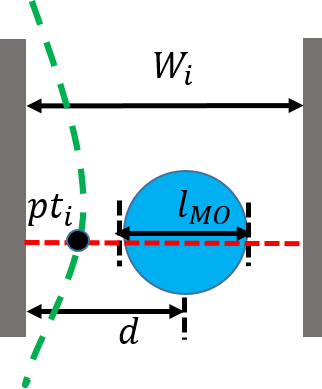}
    \caption{Blocking case. The blue circle represents the MO in a corridor with width $W_i$. The green dash curve is the planned trajectory of the robot while the red dash line is the traversal line at waypoint $pt_i$}
    \label{fig:block_anno}
\end{figure}

To calculate the blockage probability of a trajectory $T$, we take $T$ as a set of way points $pt_i, i=1,2,...,N$ and compute the blockage probability of each way point.

We model the blockage probability based on several assumptions (shown in Fig.~\ref{fig:block_anno}):
\begin{enumerate}[label=(\roman*)]
\item The width of the passage $W_i$ at $pt_i$, 
\item The radius of the robot $r$,
\item The diameter of the MO, $l_{MO}$, modeled as a Gaussian distribution $G(\mu, \sigma)$ where  $\mu$ is the average radius of the MOs and $\sigma$ is the standard deviation.
\item The distance between the MO center and the wall $d$, modeled as a uniform distribution $d \sim U(\frac{l_{MO}}{2}, W_i-\frac{l_{MO}}{2})$, which leads to $p(d|l_{MO}, W_i)=\frac{1}{W_i-l_{MO}}$.
\end{enumerate}

Given $d$, $l_{MO}$, $r$ and $W_i$, the blockage probability  is deterministic: $p(b|d,l_{MO}, W_i, r)=1$ when the size of robot is larger than the space of both sides of the MO, represented by $2r>max(d-\frac{l_{MO}}{2}, W_i-d-\frac{l_{MO}}{2})$; otherwise, $p(b|d,l_{MO}, W_i, r)=0$. 

To obtain the blockage probability conditioned only on the passage width and the robot size $p(b|W_i,r)$, we first eliminate the dependence on $d$ through marginalization:
\begin{multline}
p(b|l_{MO}, W_i,r) =  \\
\int_{l_{MO}/2}^{W_i-l_{MO}/2} p(b|d,l_{MO},W_i,r) p(d|l_{MO},W_i,r) \,d_d 
\end{multline}

After simplification, we get:

\begin{small}
\begin{equation}
{p(b|l_{MO},W_i, r)} = 
\begin{cases}
{0,}&{l_{MO}<W_i-4r} \\ 
{\frac{4r}{W_i-l_{MO}}-1,}&{W_i-4r<l_{MO}<W_i-2r} \\
{1,}&{W_i-2r<l_{MO}<W_i} \\
{0,}&{W_i<l_{MO}}
\end{cases}
\end{equation}
\end{small}
 The blockage probability in a corridor is then:
 
 \begin{equation}
\begin{split}
p(b|W_i,r) = \int p(b|l_{MO},W_i,r) p(l_{MO})  \, d_{l_{MO}}
\end{split}
\end{equation}

Considering that $l_{MO}$ satisfies a Gaussian distribution and $p(b|l_{MO},W_i,r)$ is piecewise constant, we approximate this integral by using a sampling method.


The previous $p(b|W_i,r)$ is calculated with the assumption of a MO being on the line perpendicular to the corridor passing through $pt_i$~(the red dash line in Fig.~\ref{fig:block_anno}). 
Because $W_i$ can be obtained from $pt_i$ using a ray casting algorithm, and both are independent of $r$, the blocking probability at $pt_i$ can be expressed as $p(b|pt_i,r)=p(b|W_i,r)$. 

Assuming the MO is uniformly distributed in the space with a free area $A$, the probability that the MO is in the traversal line at $pt_i$ is $p(pt_i) = \frac{W_i K}{A}$. Here $K$ is a parameter characterising the obstacle appearance probability. It's worth noting that the assumption on uniform distribution is based on the absence of prior knowledge about the object organization in the environment. However, if prior information is available, the distribution should be adjusted accordingly.

Therefore, the probability that the robot is blocked at $pt_i$ is $p(b|pt_i,r)\times p(pt_i)$. Then, for a trajectory $T$, the blockage probability $p(b|T,r)$ can be computed by: 
\begin{equation}
    p(b|T,r)=1-\prod_{pt_i}^T(1-p(b|pt_i,r)\times p(pt_i))
\end{equation}

The estimated cost of blockage when passing the invisible region is then : 
\begin{equation}
[C_{blocked}] = p(b|T,r) \times [C_{MO}]
\end{equation}

\subsection{Decision making with uncertainty interval}
\label{Sec:decision}
The decision-making module aims to compare the costs of bypassing and removal, then to choose the one with the lower cost.

With the uncertainties considered, the final cost of each option can be calculated by
\begin{align}
    [C_{by}] &= [C^{by}_{nav}]+[C^{by}_{blocked}] \\
    [C_{re}] &= [C_{MO}]+[C_{nav}^{re}]+[C^{re}_{blocked}]
\end{align}
where $[C^{by}_{blocked}]$ and $[C^{re}_{blocked}]$ are the blockage costs of bypass and removal trajectories. 


For the decision making between the cost intervals, we apply the Laplace criterion, described in~\cite{denoeux2019decision}, to compute the average utility of the consequences of each option. Assuming the cost satisfies the uniform distribution, the utility function can be expressed as:
\begin{equation}
\begin{split}
    U=\int_{min([C])}^{max([C])} x p(x) \,dx 
    =\frac{max([C])+min([C])}{2}
\end{split}
\end{equation}
where $[C]$ is either $ [C_{re}]$ or $ [C_{by}]$ to calculate the utility while $x$ and $p(x)$ are samples in the cost interval and its probability. Finally, the option with a smaller $U$ is chosen as the navigation strategy.

\section{Simulation Experiments}
\label{sec:simu}
We first conduct individual modules evaluation in simulation, and then compare our method with the state of the art before demonstrating a real robot application.

\subsection{Simulation environment}
\begin{figure}[h]
    \centering
    \includegraphics[width=0.35\textwidth]{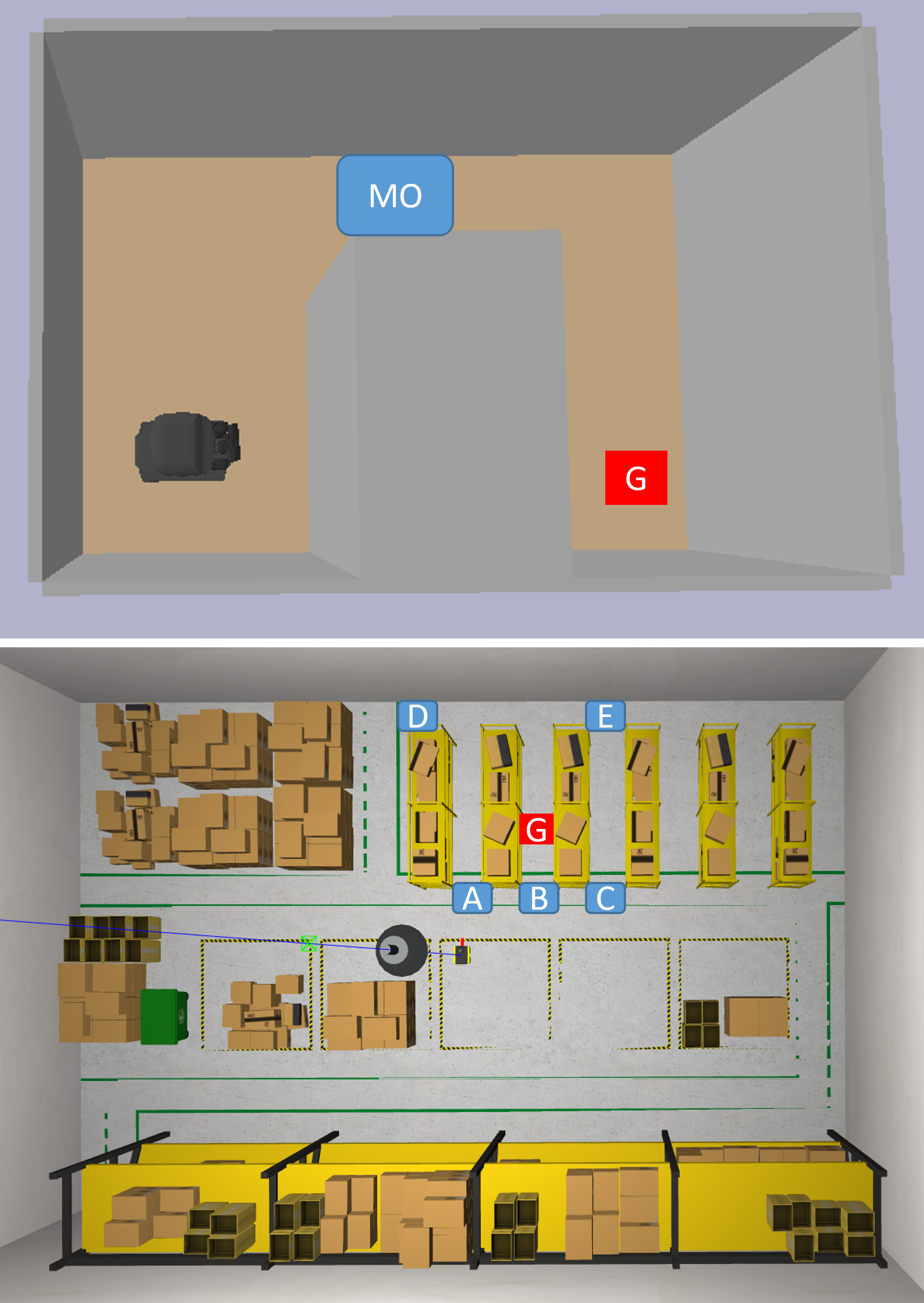}
    \caption{Simulation environments. Two environments are used including a simple room and a complex warehouse. The blue regions are possible places for the MOs and the red region is the goal.}
    \label{fig:simu_warehouse}
\end{figure}

We implement our method in two simulated environments, a simple room and a large warehouse, as shown in Fig.~\ref{fig:simu_warehouse}. The room environment is built on PyBullet~\cite{coumans2021} while the  warehouse is based on Gazebo~\cite{koenig2004design}. There is one MO in the room while multiple MOs are randomly put in the blue regions in the warehouse to simulate the variety of MO positions. A wheeled mobile robot with an arm needs to complete navigation tasks to reach the red goal.  A LiDAR and a stereo camera are used to localize the robot and detect MOs respectively. 

\subsection{Implementation details}
\label{sec:Implem}
We use ROS Noetic~\cite{ros} with Movebase as the navigation framework. 
The environment map~(including only static obstacles) is generated using GMapping~\cite{grisetti2007improved} and provided to the robot as prior knowledge. We use A*~\cite{hart1968formal} as the global path planner and TEB~\cite{rosmann2015timed} as the local path planner. The AMCL package  is employed to localize the robot from LiDAR data. 
The robot detects MOs by Aruco Marker~\cite{romero2018speeded} to reduce the classification uncertainty. 

All the experiments are implemented in Python with PyTorch~\cite{paszke2017automatic}. We use an Intel i7-12700H CPU with 16G memory for the quantitative results.

\subsection{Bypass time regression results}
\label{sec:by_predictor_eval}

\begin{figure}[h]
    \centering
    \includegraphics[width=0.35\textwidth]{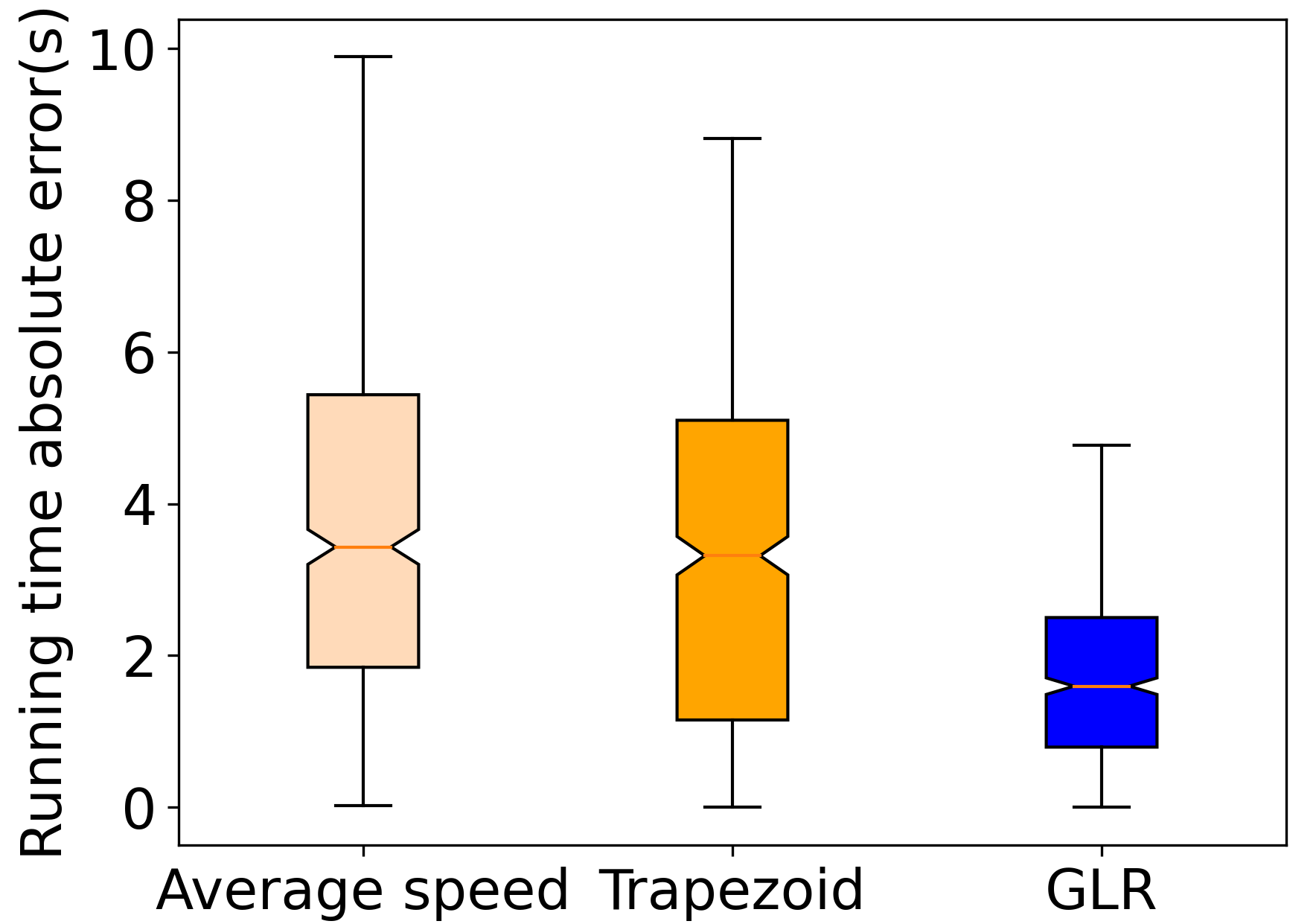}
    \caption{Boxplot of the absolute error of prediction results for the three bypass time prediction methods. The box represents the interquartile range~(IQR), with the lower and upper edges indicating the 25th~(Q1) and 75th~(Q3) percentiles,  respectively. The notch and orange line in the box marks the median value of the absolute error. Whiskers extend to the smallest and largest values within 1.5 times the IQR from Q1 and Q3.}
    \label{fig:bypass_time_pred}
\end{figure}

To demonstrate the improved bypass time prediction, we compare the GLR regression model with the average speed method from~\cite{zhang2023navigation} and a trapezoid method that considers acceleration and deceleration.  We collect a dataset manually by controlling the robot using a joystick and recording the planned path and corresponding time, including 1500 trajectory segments  as training and 600 as testing. The average speed method calculates the speed in the training dataset, then applies it in the test set. For the GLR method, the model is fitted on the training set to predict on the test set. We calculate the absolute error between the predicted and actual navigation time. The results in Fig.~\ref{fig:bypass_time_pred} show that the applied GLR method outputs more accurate estimation with the lowest median absolute error~(1.59s), compared to the average speed~(3.43s) and trapezoid methods~(3.31s). Additionally, the GLR method is more stable, with an interquartile range (IQR) of 0.69s, versus 1.35s and 1.91s for the other methods. In the case study, we found that the large error in the average speed method arises from its neglect of acceleration and deceleration during navigation. In contrast, the trapezoid method accounts for the time spent in speed changes and therefore yields smaller errors than the average speed method. However, for paths with multiple angle changes, the trapezoid method’s initial assumption, allowing only a fixed number of accelerations and decelerations, limits its adaptability. By comparison, our GLR method incorporates trajectory smoothness, enabling it to capture the key factors influencing navigation time and thus achieve the lowest prediction error.


\subsection{Action uncertainty module evaluation}
To evaluate the effectiveness of modeling action uncertainty, we compare the task completion time with (w/) and without (w/o) the action uncertainty module in two cases: one with easy MOs (90\% loading SR) and one with hard MOs (20\% SR). We test the methods in three setups: ABC, AB, BC. ABC~(resp. AB and BC) indicates three MOs are in regions A, B and C~(resp. 2 at AB and BC) in Fig.~\ref{fig:simu_warehouse}.


\begin{figure}[h]
    \centering
    \includegraphics[width=0.35\textwidth]{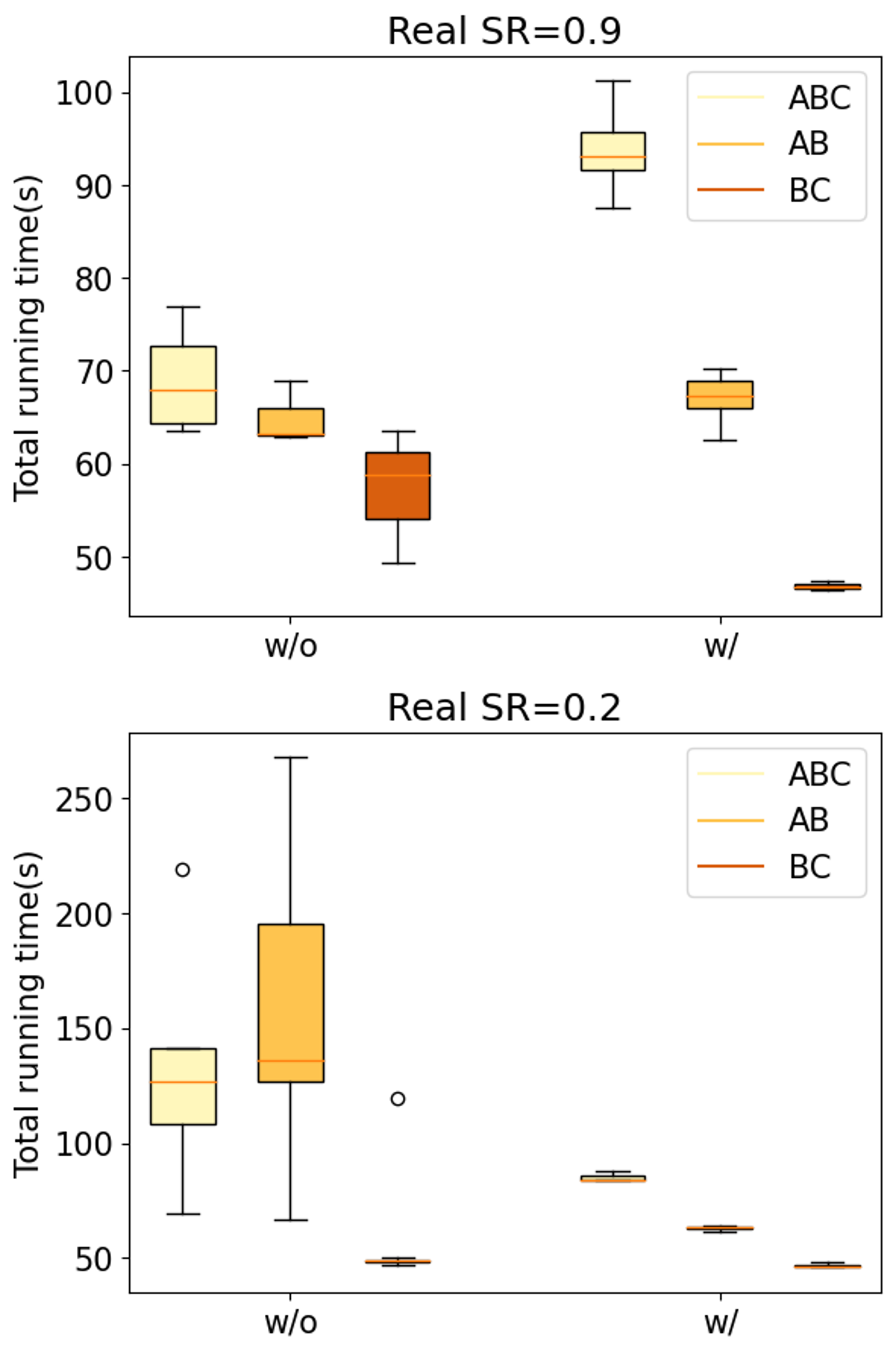}
    \caption{Task completion time comparison on methods with/without considering action uncertainty in three environments, ABC, AB, BC. The top figure shows the case with easy MO~(high SR) while the bottom one shows the hard MO~(low SR).}
    \label{fig:unc_TFF_TTF}
\end{figure}

The results in Fig.~\ref{fig:unc_TFF_TTF} show that with easy MOs (SR=90\%), the w/o method takes less time in planning and making decision as it removes the MO at B directly. Conversely, the method w/ has the robot bypass B first to check MOs in A or C. If A and C are also blocked, the robot returns and removes the MO in B, requiring more planning and navigation time.

When MOs are hard to manipulate (SR=20\%), the method w/ bypasses all MOs due to the high potential cost of the removal action, leading to faster navigation compared to w/o method that repeatedly attempts to remove MOs regardless of the cost. As for the stability, the w/ method demonstrates much lower IQR with 2.53s for easy MOs and 1.35s for hard MOs, compared to the w/o method's 6.16s and 34.14s, respectively.

\subsection{Ablation study on bias between estimated SR and real SR}
To obtain the initial SR value, we conduct prior experiments, as explained in Sec.~\ref{sec:action_unc}. However, the estimated and actual SR may differ, especially when assuming all MOs share the same SR. Since SR relates to the estimated removal cost and affects the navigation strategy, we analyze the impact of errors in estimating the SR. The results on the cases with and without estimation bias in environment ABC are shown in Table~\ref{tab:sr_bias}. All the values of running time in the table are the average of 5 trials. In the method with action uncertainty~(w/), an unbiased SR estimation gives the best navigation strategy with minimal time. Even with biased estimation, the method w/ still outperforms the method w/o, proving the effectiveness of the proposed action uncertainty module.

\begin{table}[h]
\centering
\caption{Average running time of methods with unbiased/biased estimation on SR. The cell marked bold means better result.}
\label{tab:sr_bias}

\resizebox{\columnwidth}{!}{\begin{tabular}{|c|cc|c|c|}
\hline
 & \multicolumn{1}{c|}{Estimated SR}          & Real SR & w/o (s)          & w/ (s)        \\ \hline
\multirow{4}{*}{\begin{tabular}[c]{@{}c@{}}Unbiased\end{tabular}} & \multicolumn{1}{c|}{0.90}                  & 0.90 & \textbf{68.99$\pm$4.42}   & 93.80$\pm$4.53   \\ \cline{2-5} 
 & \multicolumn{1}{c|}{0.50}                  & 0.50    & 98.31$\pm$21.81  & \textbf{94.19$\pm$12.86}  \\ \cline{2-5} 
 & \multicolumn{1}{c|}{0.20}                  & 0.20    & 164.15$\pm$80.89 & \textbf{85.14$\pm$1.92}   \\ \cline{2-5} 
 & \multicolumn{2}{c|}{Overall}                             & 110.48       & \textbf{91.04}        \\ \hline
\multirow{7}{*}{\begin{tabular}[c]{@{}c@{}}Biased \end{tabular}}   & \multicolumn{1}{c|}{\multirow{2}{*}{0.90}} & 0.20 & \textbf{164.15$\pm$80.89} & 168.19$\pm$39.50 \\ \cline{3-5} 
 & \multicolumn{1}{c|}{}                      & 0.50    & \textbf{98.31$\pm$21.81}  & 109.24$\pm$15.91 \\ \cline{2-5} 
 & \multicolumn{1}{c|}{\multirow{2}{*}{0.50}} & 0.20    & 164.15$\pm$80.89 & \textbf{113.46$\pm$22.17} \\ \cline{3-5} 
 & \multicolumn{1}{c|}{}                      & 0.90    & \textbf{68.99$\pm$4.42}   & 91.03$\pm$6.95   \\ \cline{2-5} 
 & \multicolumn{1}{c|}{\multirow{2}{*}{0.20}} & 0.50    & 98.31$\pm$21.81  & \textbf{85.57$\pm$2.31}   \\ \cline{3-5} 
 & \multicolumn{1}{c|}{}                      & 0.90    & \textbf{68.99$\pm$4.42}   & 84.45$\pm$2.42   \\ \cline{2-5} 
 & \multicolumn{2}{c|}{Overall}                             & 110.48       & \textbf{108.66 }      \\ \hline
\end{tabular}}

\end{table}

\begin{table}[h]
\centering
\caption{Average running time of method with/without blockage uncertainty in two environments}
\label{tab:blockage_abla}

\begin{tabular}{|c|cc|}
\hline
    & \multicolumn{2}{c|}{Methods}              \\ \hline
Env & \multicolumn{1}{c|}{w/o (s)}          & w/ (s)\\ \hline
AB  & \multicolumn{1}{c|}{\textbf{67.04$\pm$2.54}}   & 77.77$\pm$2.63 \\ \hline
ABE & \multicolumn{1}{c|}{141.02$\pm$16.81} & \textbf{90.08$\pm$2.25} \\ \hline \hline
Overall & \multicolumn{1}{c|}{104.03}       & \textbf{83.92}     \\ \hline
\end{tabular}

\end{table}

\begin{figure*}[ht]
    \centering
    \includegraphics[width=0.7\textwidth]{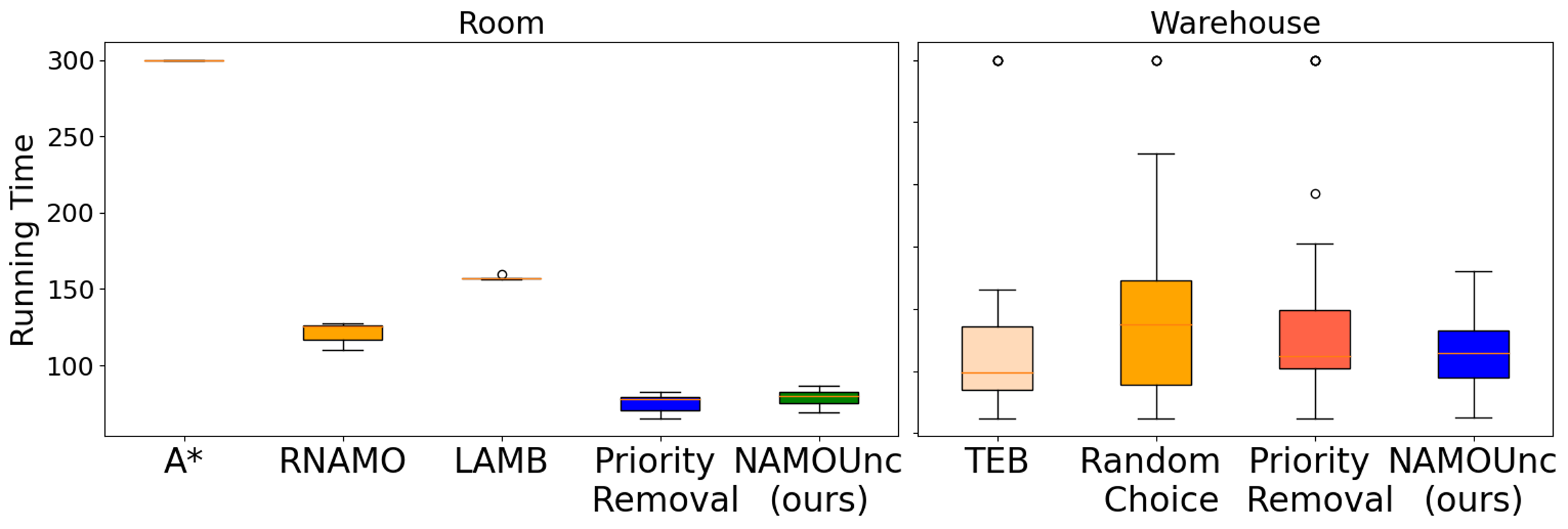}
    \caption{Running time in two simulation environments (around 70 trials for each environment), the room (left) and the warehouse (right). The outliers (circle around 300 in the figures) represent the failure cases. The orange line in the box marks the median value.}
    \label{fig:simu_res}
\end{figure*}

\subsection{Blockage uncertainty module evaluation}
To compare the impact of introducing the blockage uncertainty module, we design two environments AB and ABE for evaluation~(with obstacles at the corresponding positions shown in Fig.~\ref{fig:simu_warehouse}). Environment AB demonstrates the difference of navigation strategy due to blockage uncertainty while ABE illustrates the advantage of the blockage uncertainty module when an unexpected MO appears on the detour. 

The evaluation results in Table~\ref{tab:blockage_abla} report the mean runtime of 5 trials. In environment AB, without considering the blockage uncertainty~(w/o), the robot bypasses all the MOs. In contrast, the w/ method chooses to remove the MO in region B considering the potential blockage risk of the detour in narrow passage. In AB where no surprising MO appears, the bypass takes less time than the removal. However, in ABE, where an unexpected MO blocks the detour, the method w/o bypasses the MO in region B, and the MO in E (failing to find a suitable stock region for removing E), finally it returns to remove the MO at B, taking much longer time than the method w/ that removes B initially. From the overall performance, the method w/ is more efficient comparing to the method w/o.

\subsection{Overall comparison}

We evaluate the overall performance in the two simulation environments~(Fig.~\ref{fig:simu_warehouse}), with configurations of ABC, AB, BC, ABE, ABD, BCE in the warehouse. The starting points and goal points are randomly selected to create different navigation tasks. Then, we compare our method with some baseline methods, priority bypass~\cite{rosmann2015timed,hart1968formal}, priority removal, random choice method and LaMB~\cite{muguira2023visibility}. 

The priority bypass methods including TEB~\cite{rosmann2015timed} and A*~\cite{hart1968formal}, bypass all the obstacles but fail if there is no alternative way to the goal. The priority removal method refers to a category of NAMO methods~\cite{ellis2022navigation,wang2020affordance}, which removes the MOs that block the path without considering the removal cost. The random choice method chooses to remove or bypass with a probability~(0.5 in our experiments). The LaMB method~\cite{muguira2023visibility} is one of the latest methods considering the partial observation constraints in NAMO tasks but limited to small scale environments.

We record the running time of each method. Tasks are marked as failed if the goal is not reached within 300 seconds. Results are shown in Fig.~\ref{fig:simu_res}. In the room environment, the only path to the goal is blocked by a MO. Therefore, the bypass method (A*) always fails. The NAMOUnc and priority removal methods complete the task more quickly, with the priority removal method slightly faster (77.68s vs. 79.56s) since it takes less time to plan bypass and make decision.
In the warehouse environment, the NAMOUnc and the TEB methods finish the task with comparable time cost but NAMOUnc has no failures. 

\section{Real Experiments}
\label{sec:real}
\subsection{Environment description}
\begin{figure}[h]
    \centering
    \includegraphics[width=0.45\textwidth]{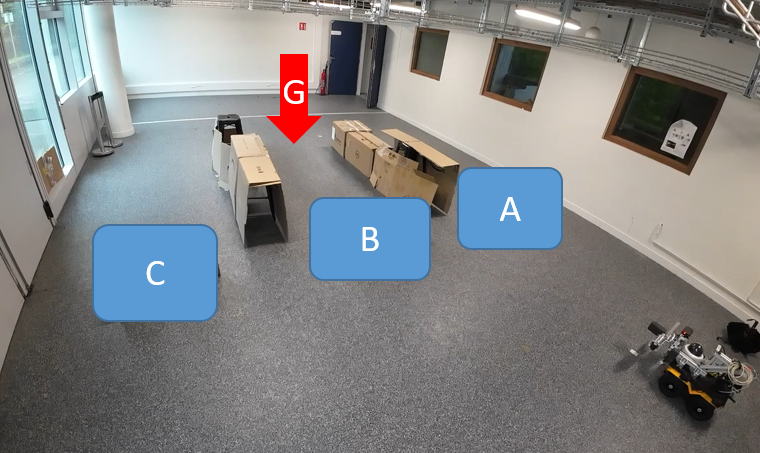}
    \caption{Real environment setup. The blue regions marked as A, B, C are possible positions of MOs. The red arrow with G is the goal.}
    \label{fig:real_env}
\end{figure}

To evaluate the performance of our method in real applications, we use a Jackal differential mobile platform in a small warehouse-like environment. As shown in Fig.~\ref{fig:real_env}, there are maximum 3 MOs and the robot should navigate to a goal G. It is equipped with a LiDAR and a RealSense camera to observe the environment, an arm to lift MOs and the control software described in Sec.~\ref{sec:Implem}.

\begin{table*}[ht]
\centering
\caption{Overall performance comparison in the real environment. The columns marked bold represent the best results.}
\label{tab:real_overral}
\begin{tabular}{|c|c|c|c|c|}
\hline
 Env   & TEB\cite{rosmann2015timed}        & \begin{tabular}[c]{@{}c@{}}Priority \\ Removal\end{tabular} & \begin{tabular}[c]{@{}c@{}}Random \\ Choice\end{tabular} & \begin{tabular}[c]{@{}c@{}}NAMOUnc\\(Ours) \end{tabular}       \\ \hline
ABC & 300.00$\pm$0.00    & \textbf{96.33$\pm$9.01} & 115.34$\pm$14.58  & 137.69$\pm$13.19 \\ \hline
BC  & \textbf{46.66$\pm$4.56} & 94.78$\pm$5.08 & 72.73$\pm$19.94   & 50.41$\pm$2.62   \\ \hline
AB  & \textbf{61.11$\pm$4.90} & 97.94$\pm$5.72 & 107.44$\pm$22.40  & 72.56$\pm$2.09   \\ \hline \hline 
Overall &   135.93         & 96.35      & 98.5          & \textbf{86.88}        \\ \hline
\end{tabular}

\end{table*}

\subsection{Experiment results}

We randomly pick goal points to create different NAMO tasks and record the running time with different MO setups, including environments ABC (all paths to the goal are blocked), AB and BC (at least one path to the goal is feasible). The quantitative results are shown in Table~\ref{tab:real_overral}, where each cell indicates the average running time and corresponding standard deviation on 5 repeated trials. The priority bypass method (TEB) fails in the ABC environment because it finds no feasible path to the goal. Although the proposed NAMOUnc method does not achieve the best result in every individual environment, it attains the best overall performance across the three setups in terms of average running time. This shows that it achieves a good trade-off between completeness and efficiency in the search for a solution.


\section{CONCLUSION AND FUTURE WORK}
\label{sec:con}
We have presented a NAMO framework capable of planning task and motion under four kinds of uncertainties related to observation, action, model and observability. 
Our planner jointly optimizes success rate and running time. Experimental results in both simulation and real environments demonstrate its ability to balance these objectives, suggesting potential extensions to optimize additional objectives like energy and safety.

A relevant perspective is to overcome simplified assumptions. For instance, observation noise may not always follow a Gaussian distribution, and action failures can stem from various factors, such as mechanical constraints that can depend on the obstacles and the environment. We believe that with more data collection, a more accurate model can be developed, such as a planner based on large language models~\cite{honerkamp2024language}, enabling the proposed method to provide more robust and effective solutions for NAMO tasks.

\section*{\uppercase{Acknowledgements}}
The publication of this research was supported by the National Natural Science Foundation of China [Grant 42101445] and the Director Foundation of Guangdong Laboratory of Artificial Intelligence and Digital Economy(SZ) [Grant 25420001 and 24420004].

\bibliographystyle{apalike}
{\small
\bibliography{example}}

\end{document}